\documentclass{article}
\usepackage{ijcai21}

\usepackage{times}
\usepackage{soul}
\usepackage{url}
\usepackage[hidelinks]{hyperref}
\usepackage[utf8]{inputenc}
\usepackage[small]{caption}
\usepackage{graphicx}
\usepackage{amsmath}
\usepackage{amsthm}
\usepackage{booktabs}
\usepackage{algorithm}
\usepackage{algorithmic}
\usepackage{amsfonts}
\usepackage{multirow}
\usepackage{stfloats}

\usepackage{array}
\newcommand{\tabincell}[2]{\begin{tabular}{@{}#1@{}}#2\end{tabular}}

\title{Automatic Learning to Detect Concept Drift}

\author{
Hang Yu$^1$
\and
Tianyu Liu$^1$\and
Jie~Lu$^{1}$\And
Guangquan Zhang$^1$
\affiliations
$^1$Australian Artificial Intelligence Institute, University of Technology Sydney. Australia.
\emails
\{hang.yu, Tianyu.Liu-1\}@student.uts.edu.au,
\{Jie.Lu, guangquan.zhang\}@uts.edu.au
}

\begin{document}

\maketitle

\begin{abstract}
Many methods have been proposed to detect concept drift, i.e., the change in the distribution of streaming data, due to concept drift causes a decrease in the prediction accuracy of algorithms. However, the most of current detection methods are based on the assessment of the degree of change in the data distribution, cannot identify the type of concept drift. In this paper, we propose Active Drift Detection with Meta learning (Meta-ADD), a novel framework that learns to classify concept drift by tracking the changed pattern of error rates. Specifically, in the training phase, we extract meta-features based on the error rates of various concept drift, after which a meta-detector is developed via a prototypical neural network by representing various concept drift classes as corresponding prototypes. In the detection phase, the learned meta-detector is fine-tuned to adapt to the corresponding data stream via stream-based active learning. Hence, Meta-ADD uses machine learning to learn to detect concept drifts and identify their types automatically, which can directly support drift understand. The experiment results verify the effectiveness of Meta-ADD.
\end{abstract}

\section{Introduction}

\begin{figure}
\centering
\includegraphics[width=.35\textwidth]{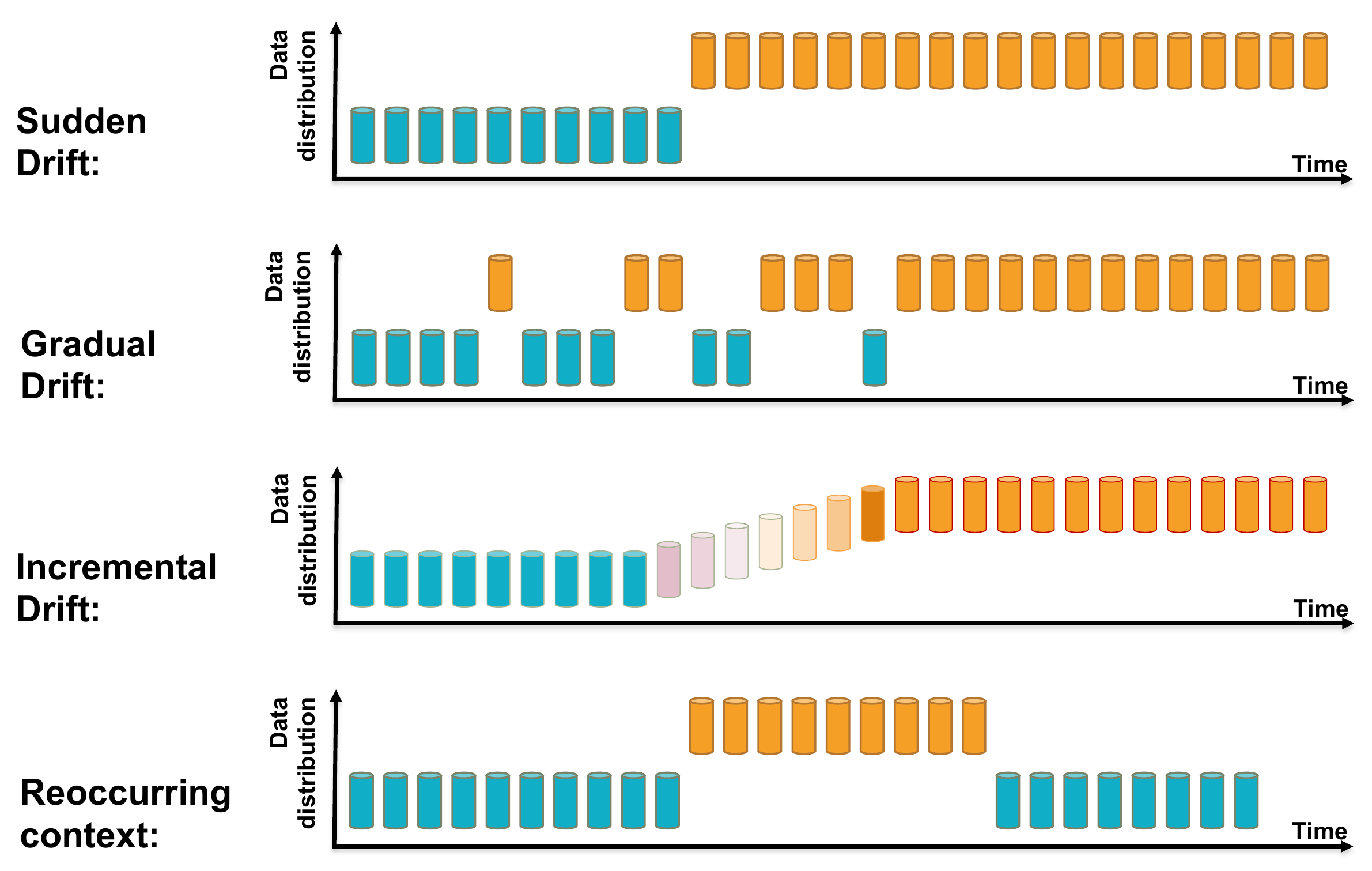}
\caption{Types of concept drift.}
\label{fig1}
\end{figure}

Concept drift is a hot research topic in data stream mining \cite{he2019online}, incremental learning \cite{losing2018incremental}, nonstationary learning \cite{pratama2019deep}, and it can be further extended to machine learning. In the machine learning area, concept drift is a phenomenon in which the statistical properties of the predicted variable change over time in an arbitrary way \cite{lu2017dynamic}. This causes the predictions to become less accurate as time passes, and therefore needs to be detected. Figure \ref{fig1} shows the four types of concept drift, categorized according to the changed pattern of data distribution \cite{gama2014survey}.

In the two recent decades, although many methods of detecting concept drift have been proposed, they all, in general, follow a framework \cite{8496795} which includes four steps: 1) data retrieval aims to retrieve data chunks from data streams; 2) data modeling aims to abstract the retrieved data and extract the key features; 3) test statistics calculation is the measurement of dissimilarity, or distance estimation \cite{de2018concept}; 4) hypothesis testing uses a specific hypothesis test to evaluate the statistical significance of the change observed in step 3, or the p-value. 

However, this framework faces two drawbacks: the first is the cold start problem. In current drift detection methods, an initial window is needed to collect the basic statistic properties for the hypothesis test. As a result, no detection strategies can be implemented in the initial window, but there may be concept drift in the initial window. The second is, to the best of our knowledge, the existing methods can only identify whether concept drift has occurred, but they are unable to identify what type of concept drift has occurred. The reason for this is that although the change of data distribution can be represented by test statistics such as average error rate, test statistics cannot represent the changed pattern of data distribution, i.e., the relationship between concept drift at adjacent timestamps cannot be captured. In practice, knowing the type of concept drift is useful to understand them. For example, if a person's weight suddenly decreases, this may indicate he has health problems which should be addressed instantly, but if a person's weight gradual decreases over a long time, such as over one year, the situation may be normal because the man is making an effort to lose weight.

In this paper, we propose a novel framework to learn a meta-detector with the ability to classify concept drift which can be trained in advance. When a data stream is inputted, it can be used to directly classify concept drift without setting an initial window. However, this framework faces two main challenges. On the one hand, although concept drift can be classified into four types, the presentations, i.e., changed pattern of data distribution, of one type of drifts is infinite possibly over 1000000. By contrast, the presentations of this type of drifts, which are included in training data, are very small possibly 1000. On the other hand, for each type of concept drift, each data stream may have its own unique presentation, so the meta-detector must have the ability to be able to adjust to fit the detected data stream. To address these two challenges, we propose Active Drift Detection with Meta learning (Meta-ADD). The contributions of this paper are summarized as follows:

\begin{itemize}
\item We propose a framework that learns to detect drift by learning the presentation of each type of concept drift. As a result, in our framework, not only can concept drift be detected, the type of drift can also be identified to help us understand them.
\item We propose to transform the pre-training of a meta-detector into a few-shot learning problem, and thereby proposing a meta-learning method \cite{finn2017model} based on prototypical-networks \cite{snell2017prototypical} to learn the meta-detector. In this meta-detector, various concept drift classes can be represented as a corresponding single prototype. 
\item To improve the accuracy of detecting concept drift, we propose a stream-based active learning algorithm (SAL) which is capable of handling various concept drifts by adapting the meta-detector to the underlying distribution in the stream.
\end{itemize}

The rest of this paper is organized as follows. Section \ref{sec2} presents the related work. Section \ref{sec3} provides the details of our proposed Meta-ADD framework. Section \ref{sec4} discusses the experiment results for several well-known datasets in this research area. Finally, Section 5 concludes the paper and presents the future work.

\section{Related work}\label{sec2}

The survey of the literature related to drift detection methods reveals that in general, these can be divided into three categories:

\subsection{Error Rate-based Drift Detection}

Error rate-based drift detection methods form the largest category of drift detection algorithms. The intuition behind this type of algorithms is to monitor fluctuations in the error rate as time passes. Once the change of the error rate is proven to be statistically significant, a drift alarm will be triggered. One of the top referenced this type of algorithms is the Drift Detection Method (DDM) \cite{gama2004learning} which was the first algorithm to contain defined warning and drift levels for signalling concept drift. Many subsequent algorithms have adopted a similar implementation, e.g., Early Drift Detection Method (EDDM) \cite{baena2006early}, Heoffding's inequality-based Drift Detection Method (HDDM) \cite{6871418} and based on the Kolmogorov-Smirnov (KS) statistical test. In contrast to methods such as DDM, a two-time window-based drift detection algorithm called Adaptive Windowing (ADWIN) \cite{bifet2007learning} was proposed. In ADWIN, the size of the compared windows can be adjusted automatically.

\subsection{Data Distribution-based Drift Detection}

The second largest category of drift detection algorithms is data distribution-based drift detection. In this category of algorithms, the dissimilarity between the distribution of new data and historical data is quantified by using a distance metric. Once the dissimilarity is proven to be statistically significantly different, a drift alarm will be triggered. According to the literature, the first formal treatment of change detection in data streams was proposed by \cite{kifer2004detecting}. Another typical data distribution-based drift detection algorithm is the Information-Theoretic Approach (ITA) \cite{dasu2006information}. Similar implementations have been adopted in the Competence Model-based drift detection (CM) \cite{lu2014concept}, Equal Density Estimation (EDE) \cite{8897716}, and Local Drift Degree-based Density Synchronized Drift Adaptation (LDD-DSDA) \cite{liu2017regional}.

\subsection{Multiple Hypothesis Test Drift Detection}

Multiple hypothesis test drift detection algorithms apply similar techniques to those mentioned in the previous two categories. The novelty of this type of algorithms is that the multiple hypothesis tests is used to detect concept drift in different ways. For example, the Just-In-Time adaptive classifier (JIT) \cite{alippi2008just} is the first algorithm to set multiple drift detection hypotheses in this way, but Hierarchical Change-Detection Tests (HCDTs) \cite{7386680} is the first attempt to address concept drift using a hierarchical architecture.

\section{Active Drift Detection with Meta-learning}\label{sec3}

In this section, we elaborate on the Active Drift Detection with Meta learning (Meta-ADD). Meta-ADD includes two main phases: training phase and detection phase. During the training phase, we extract the meta-features of various concept drifts, and then learn a meta-detector where various concept drift classes can be represented as a corresponding single prototype \cite{8103824}. In the detection phase, the learned meta-detector is used to detect concept drift and is further fine-tuned via stream-based active learning \cite{krawczyk2019adaptive}. 


\subsection{Extracting Meta-Features}

In this subsection, we describe the extraction of the meta-features that can be used across different datasets. Although there are four types of concept drift, the reoccurring concept type focuses on tracking whether an old concept occurs again. In contrast, our work focuses on detecting concept drift immediately instead of considering whether a concept is an old concept. So we did not use the reoccurring concept type and instead used a concept type we named normal to indicate that there was no concept drift. It is proved that a change in the data distribution can be represented by a change of error rates, so intuitively, the change of error rate is critical for deciding which type of concept drift has occurred:

\begin{itemize}
\item {\bf Sudden drift:} the error rate will decrease sharply within a short time.
\item {\bf Gradual drift:} the error rate will fluctuate over a period of time.
\item {\bf Incremental drift:} the error rate will increase incrementally over a period of time.
\item {\bf Normal:} the error rate will remain stable.
\end{itemize}

Based on the intuitions above, we empirically extract the gap between the average error rates of two windows as meta-features. Specifically, for each type of concept drift, we first generate $N$ number of data streams. A prediction model such as Neural network, decision tree or SVR was directly applied to a data stream to obtain error rate $e_t$ at each timestamp $t$. In addition, we use a window $W_i$ with $n$ length, that is, there are $n$ number of samples in $W_i$, so the average error rates of $W_i$ are calculated as:

\begin{equation}
\hat{e}_{i}=\frac{\sum_{t=1}^{n} e_{i}}{n}
\end{equation}

Furthermore, the gap between the average error rates of two windows is calculated as: 

\begin{equation}
G a p_{i}=\hat{e}_{i+1}-\hat{e}_{i}
\end{equation}

Assume there are $m=l \times n$ samples in each data stream, so $l$ number of windows can be obtained, and thereby $l-1$ is the number of $Gap_i$ which can be obtained. Hence, a training sample in meta-training is represented as $G_{i}:\{\hat{X} \times \hat{y}\} \in \mathbb{R}^{l}$, where $\hat{X}=\left\{G a p_{1}, \ldots, G a p_{l-1}\right\}$, and $\hat{y} \in 1, \ldots, K$ is the corresponding labels, and thereby an original dataset with $N$ number of data streams, i.e., $g:\{X \times y\}^{N \times m}$, is mapped to $G:\{\hat{X} \times \hat{y}\} \in \mathbb{R}^{N \times l}$, such that $G$ is less dependent on the dataset. 

Note that our framework allows flexibility in the choice of features. For example, we may be able to improve the performance by using different predicted models or a more intelligent windowing strategy. To make our contribution focused, we adopt simple models and tumbling windows in all our experiments, which results in a reasonable performance based on our empirical results. How we can better model the transferable information will be interesting future work to enhance the Meta-ADD framework.

\subsection{Meta-detector via a Prototypical Neural Network}

In this subsection, a prototypical neural network is applied to extract meta features from various data streams. The concept drift detection issue is considered to be a classification task by learning a neural network for mapping the data stream into an embedding space and extracting prototypes for various concept drift categories to be the mean vector of the embedded support data stream. The detection of concept drift is performed for the embedded data stream by selecting the nearest concept drift prototype.

\begin{figure*}
\centering
\includegraphics[width=1\textwidth]{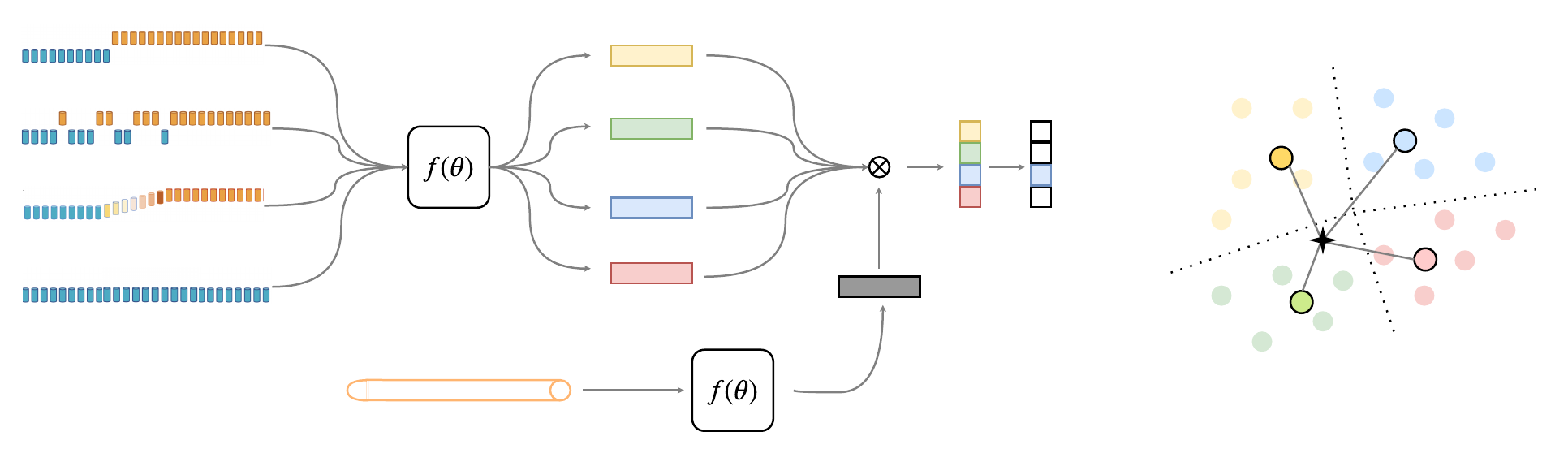}
\caption{The prototypical neural network. (The left side is the whole structure of prototypical neural network, which consists of embedding and clustering phase; in the right side the new sample is classified by finding the nearest prototype over a distance in the embedding space)}
\label{fig2}
\end{figure*}

Based on the support dataset $G$, the prototypical neural network consists of two steps, as illustrated in Figure \ref{fig2}. First, an {em embedding neural network} $f_{\theta}: \mathbb{R}^{l} \rightarrow \mathbb{R}^{M}$ with the learnable parameter $\theta$ maps the stream data $X_i$ into the embedding space $f_{\theta}\left(X_{i}\right)$, then a single prototype representation $c_k$ for each class $k$ can be taken as the mean of its support set in the embedding space:

$$
c_{k}=\frac{1}{\left|S_{k}\right|} \sum_{\left(X_{i}, y_{i}\right) \in R} f_{\theta}\left(\hat{X}_{i}\right)
$$
where $S_k$ is the set of stream data with label $k$. Given a distance function $d: \mathbb{R}^{M} \times \mathbb{R}^{M} \rightarrow \mathbb{R}$, eg. cosine similarity, classification is then performed for a query point $X$ by finding the nearest prototype $c_k$ over a distance in the embedding space:

$$
p_{\theta}(\hat{y}=k \mid \hat{x})=\frac{\exp \left(-d\left(f_{\theta}(\hat{x}), c_{k}\right)\right)}{\sum_{k^{\prime}} \exp \left(-d\left(f_{\theta}(\hat{x}), c_{k^{\prime}}\right)\right)}
$$

We then use negative log-likelihood function $J(\theta)=-\log p_{\theta}(y=k \mid \hat{x})$ as the training objective. During the training phase, $N_s$ examples are selected randomly for each concept drift class to represent the prototype in the embedding space, then $N_q$ samples are chosen randomly as the query set to measure the estimation accuracy of the learned meta learner. The pseudocode to train the model is provided in Algorithm \ref{alg1}. 

\begin{algorithm}[tb]
\caption{training meta-detector}
\label{alg1}
\textbf{Input}: training data steam $G:\left\{\left(X_{1}, y_{1}\right), \ldots,\left(X_{l}, y_{l}\right)\right\}$, where the corresponding label $y_{i} \in 1, \ldots, K$, $D_k$ is the set of stream data with label $k$.\\
\textbf{Parameter}: the number of support data stream $N_s$, the number of query data stream $N_q$, the number of concept drift class $K$. \\
\textbf{Output}: the training loss of $J(\theta)$\\
1: {\bf for} $k$ in $\{1, \ldots, K\}$\\
2:\quad $S_{k} \leftarrow$ RAMDOMSAMPLE $\left(D_{k}, N_{s}\right)$\\
3:\quad $Q_{k} \leftarrow$ RANDOMSAMPLE $\left(D_{k}, N_{a}\right)$\\
4:\quad $c_{k}=\frac{1}{N_{s}} \sum_{\left(X_{i}, y_{i}\right)} f_{\theta}\left(X_{i}\right)$\\
5: {\bf end for}\\
6: $J \leftarrow 0$\\
7: {\bf for} $k$ in $\left\{1, \ldots, N_{q}\right\}$\\
8:\quad {\bf for} $(x, y)$ in $Q_{k}$\\
8:\qquad $J=J-\frac{1}{K N_{q}} \log \frac{\exp \left(-d\left(f_{\theta}({x}), c_{k}\right)\right)}{\sum_{k^{\prime}} \exp \left(-d\left(f_{\theta}({x}), c_{k^{\prime}}\right)\right)}$\\
10:\quad {\bf end for}\\
11: {\bf end for}
\end{algorithm}

\subsection{Active Drift Detection}

Although a meta-detector can be obtained in the training process and can be directly applied to detect concept drift, the representation of each type of concept drift on different data streams can vary and each has unique characteristics. Hence, when the meta-detector is applied to a data stream, it must be able to adapt to this data stream to achieve better performance. 

To update the meta-detector, when we collect a sample $S_{i}=\left\{e_{i-i}, \ldots e_{i}\right\}$ where $e_i$ is the obtained error rate at timestamp $i$, we need to know the true type of concept drift which has occurred on this sample to update the meta-detector. However, unfortunately, the true type of concept drift is unknown. Although the true type can be inputted manually (label cost), the label cost is too expensive if the true label is manually input as soon as sample $S_i$ is collected. Hence, only the most valuable samples for labeling to update the meta-detector are used.

In Meta-ADD, the meta-detector denotes the probability $P(c)$ that sample $S_i$ belongs to class $c$, where $\sum_{c \in C} P(c)=1$. Then, the entropy of classification can be calculated as:

\begin{equation}\tag{1}
H(C)=-\sum_{c \in C} P(c) \log p(c)
\end{equation}

The entropy means the uncertainty of classification. A larger entropy means the type of this concept drift is more uncertain, and therefore this type of concept drift needs to be labelled manually.

\begin{figure}[t]
	\centering
	\includegraphics[width=0.5\textwidth]{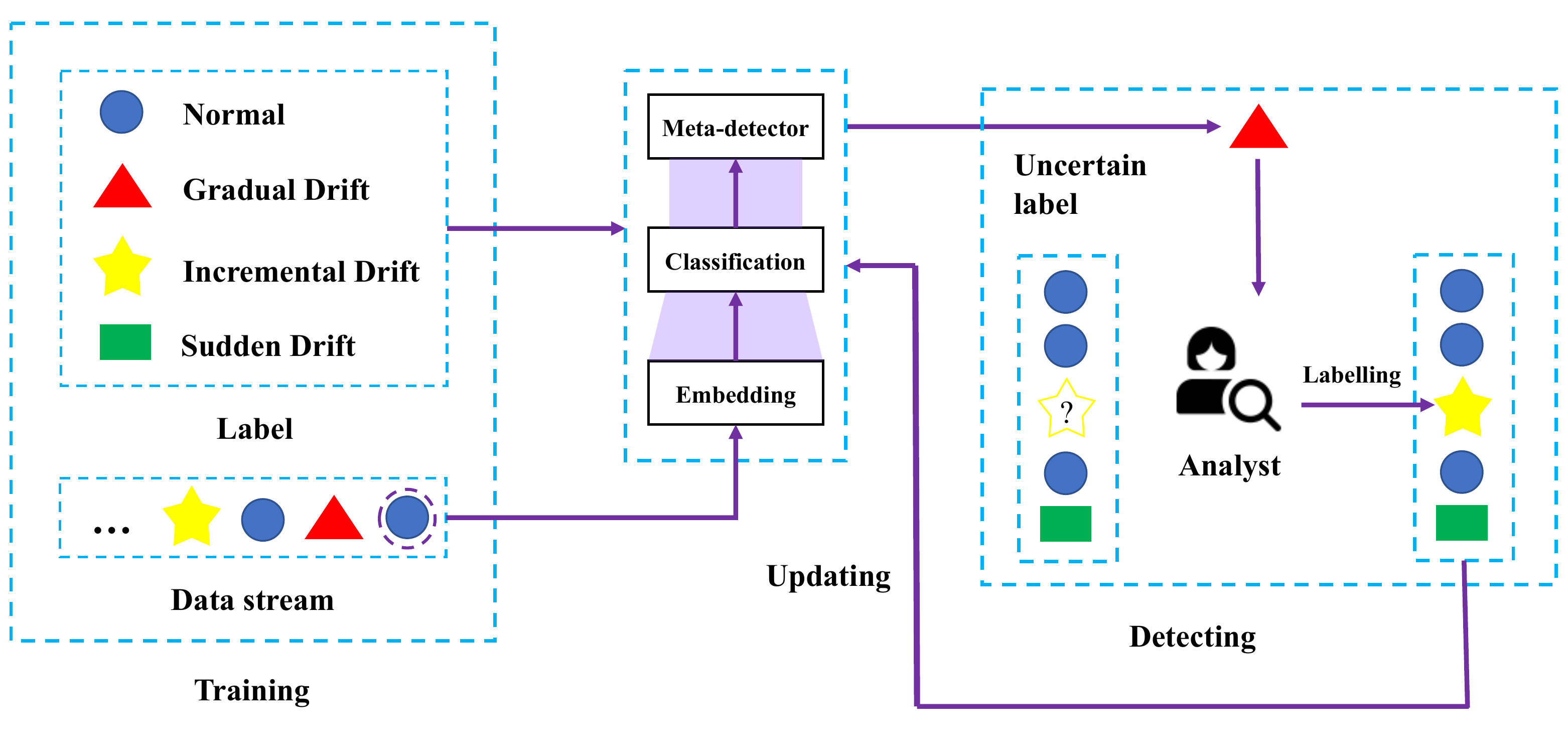}
	\caption{Structure of Meta-ADD algorithm.}
	\label{fig3}
\end{figure}

The structure of the proposed Meta-ADD algorithm is shown in Figure \ref{fig3}.

\section{Experiments}\label{sec4}

To evaluate the Meta-ADD, we conducted four experiments. The first two experiments described in Section \ref{sec41} were designed to analyze the effectiveness of Meta-ADD in identifying the type of concept drift which has occurred. The third experiment described in Section \ref{sec42} compares the difference in the performance of other drift detection algorithms and Meta-ADD on toy datasets with known drifts. The fourth experiment described in Section \ref{sec43} evaluates the strengths and weaknesses of Meta-ADD on real-world datasets with unknown drifts. All experiments are implemented in Python 3.7 version on Windows 10 running on a PC with system configuration Intel Core i7 processor (2.80 GHz) with 16-GB RAM.

\subsection{Demonstration of Meta-ADD}\label{sec41}

\subsubsection{Experiment 1. Drift detection on data streams with different meta-learning structures}\label{sec411}

In this experiment, we focus on investigating how meta-learning impacts learning performance, because, in our proposed Meta-ADD, the quality of the detector depends on the results of meta-learning, and the results of meta-learning depend on the structure of prototypical-networks. Hence, the intuition is to generate a set of data streams with predefined types of concept drift, and then observe the results when Meta-ADD with different types of prototypical-networks and parameters is applied to those streams. We expect to see that meta-learning enables our detector to classify concept drift.

{\bf Experiment Settings.} First, based on Python package skmultiflow \cite{montiel2018scikit} (version 0.5.3), we generate a total of 8000 data streams, and each data stream has 2000 instances to simulate the prequential prediction errors of a learning model. Furthermore, each data stream only has one type of drift. Then, we select 1250 data streams from each type of concept drift, i.e., a total of 5000 data streams, as an original training dataset to extract meta-features. The remaining data streams are mixed as testing datasets. GaussianNB is imported from sklearn \cite{pedregosa2011scikit} as the default learning model because it has a fast-processing speed and a partial fit option. As for the structures of the prototypical neural network, two embedding functions: a 4-layered fully connected neural network and recurrent neural network are selected as the nonlinear mapping. In the training phase, for each concept drift category, 5 support data streams are chosen randomly to act as the prototype and 15 query data streams are selected to evaluate the performance of the model. During the testing phase, 20 support data streams are randomly selected from the training dataset to perform as the final prototype. For all the experiments, ReLU activations are applied and the distance function is the cosine similarity function. All our models are trained via SGD with Adam, with a learning rate of $10^{-3}$.

\begin{table}[h]
	\small
	\centering
	\begin{tabular}{ccccc}
		\toprule
		Type & Network & Acc \\
		\midrule
		\multirow{2}*{Sudden}
		&{FCN} &98.5\%\\
		&{RNN}      &32.3\%\\\midrule
		\multirow{2}*{Gradual}
		&{FCN} &94.2\%\\
		&{RNN}      &53.4\%\\\midrule
		\multirow{2}*{Incremental}
		&{FCN} &91.3\%\\
		&{RNN}      &52.1\%\\\midrule
		\multirow{2}*{Normal}
		&{FCN} &93.8\%\\
		&{RNN}      &82.2\%\\
		\bottomrule
	\end{tabular}
	\caption{Accuracy of different deep structures. (Acc is the abbreviation of accuracy; 50-25 means the number of meta-features is 50 and the sampling frequency is 25.)}
	\label{tab1}
\end{table}


{\bf Findings and Discussion.} The results of Experiment 1 are shown in Table \ref{tab1}. From Table \ref{tab1}, we can see Meta-ADD was able to identify the simulated concept drifts and, overall, the results were in line with expectations, in that the FCN achieves the better performance because the relationship between samples in a data stream is not required strong. In addition, we can see each type of concept drift, with same settings, performed differently. 


\subsubsection{Experiment 2. Drift Detection on data streams with different meta-features}\label{sec412}

In this experiment, we focus on investigating how the meta-features impact learning performance. However, in our proposed Meta-ADD, extracting meta-features depends on two parameters $l$ and $n$, where $l$ means the number of meta-features, $n$ means the sampling frequency. Hence, the intuition is to generate a set of data streams with a predefined type of concept drift, and then observe the results when Meta-ADD with different $l$ and $n$ is applied to these streams. We expect to see Meta-ADD with different $l$ and $n$ perform differently on classifying concept drifts. 

{\bf Experiment Settings.} We again chose GaussianNB as the default learning model and the same parameter settings as in Experiment 1. Based on the results of experiment 1, the deep structure of meta-learning is implemented on FCN. However, to explore the influence of $l$ and $n$ when extracting the meta-features, the number of meta-features $l$ was changed within the set $l_{{vaild }} \in\{50,100,200\}$, and the sampling frequency $n$ was also changed within the set $n_{{vaild }} \in\{1,25,50\}$.

\begin{table}[ht]
\small
\centering
\begin{tabular}{cccc}
\toprule
&$n=1$	&$n=25$	&$n=50$\\
\midrule
$l = 50$	&71.92\%&	86.45\%	&92.57\%\\\midrule
$l = 100$	&73.83\%&	72.53\%	&94.56\%\\\midrule
$l = 200$	&70.42\%&	72.98\%	&90.68\%\\
\bottomrule
\end{tabular}
\caption{Accuracy of different meta-features}
\label{tab2}
\end{table}

{\bf Findings and Discussion.} The results of Experiment 2 are plotted in Table \ref{tab2}. From Table \ref{tab2}, two interesting points are worth mentioning: first is we can see that $l$ and $n$ influence the prediction accuracy. However, the relationship between the accuracy of the drift detection, the number of meta-features $l$, and the sampling frequency $n$, respectively, is not the same. When the $n$ increases, the prediction accuracy increases in general. In contrast, when the $l$ increases, the prediction accuracy may decreases. 

\subsection{Stream Learning on Toy Datasets with Simulated Concept Drift}\label{sec42}

\subsubsection{Experiment 3. Drift Detection on Toy Datasets}

In this experiment, we focus on investigating the performance of Meta-ADD compared to current drift detection methods. However, current drift detection methods cannot classify concept drift, so we only compare the prediction error based on extreme cases. We generate a set of data streams with a predefined type of drift, drift frequencies and severities to simulate the prequential prediction errors of a learning model with different drift detection methods. If there are less prediction errors, this means that the performance of the drift detection method is better. We expect to see that the learning model with our proposed meta-ADD achieves the best performance on different data streams with different concept drifts.

{\bf Experiment Settings.} GaussianNB is again choses as the default learning model. The toy datasets were generated by skmultiflow's built-in data generation functions. We chose SEA generator, Rotating Hyperplane generator, AGR (AGRAWAL) generator, RTG (Random Tree Generator) and RBF generator. All generators generate 10000 samples. The compared algorithms were ADWIN, DDM, EDDM, HDDM-A, HDDM-W, Page-Hinkley, KSWIN implemented in skmultiflow with their advised drift thresholds. We use the parameters with the best performance in experiment 1 to train the model and set $l$ to 100 and $n$ to 1 to extract the meta-features. 

\begin{table*}[ht]
	\small
	\centering
	\begin{tabular}{lcccccccccc}
		\toprule
		&\multicolumn{2}{c}{SEA}&\multicolumn{2}{c}{HYP}&\multicolumn{2}{c}{AGR}&\multicolumn{2}{c}{RBF}&\multicolumn{2}{c}{RTG}\\
		\toprule
		&Acc&	F1&	Acc&	F1	&Acc&	F1&	Acc	&F1	&Acc	&F1\\\toprule
		* &83.7\%	&Nan	&81.6\%&	Nan	&70.3\%&	Nan&	69.3\%&	Nan	&64.5\%&	Nan\\\midrule
		*+ADWIN	&84.3\%&	0.62	&81.6\%&	0.00&	71.3\%&	0.50&	69.3\%&	0.00&	64.5\%	&0.00\\\midrule
		*+DDM&	83.7\%&	0.00	&86.1\%	&0.63&	71.2\%&	0.50&	71.5\%&	0.66	&64.5\%	&0.00\\\midrule
		*+EDDM	&84.0\%&	0.38&	86.1\%&	0.63&	65.1\%&	0.25&	67.8\%&	0.38&	65.2\%&	0.38\\\midrule
		*+HDDM-A&	84.7\%&	0.69&	81.6\%&	0.00&	72.3\%&	0.63&	69.3\%	&0.00&	64.5\%&	0.00\\\midrule
		*+HDDM-W&	84.5\%&	0.63&	86.4\%&	0.70&	72.4\%&	0.63&	69.8\%	&0.50&	66.3\%&	0.55\\\midrule
		*+Page-Hinkley	&85.0\%&	0.71&	81.6\%	&0.00&	71.4\%&	0.50&	69.3\%&	0.00	&64.5\%&	0.00\\\midrule
		*+KSWIN	&84.1\%	&0.38&	86.2\%&	0.66&	71.5\%&	0.50&	69.9\%&	0.63&	65.2\%&	0.38\\\midrule
		*+Meta-DD	&86.1\%&	0.77&	87.8\%&	0.83&	73.2\%&	0.69&	72.2\%&	0.71&	67.4\%&	0.70\\\midrule
		*+Meta-ADD&	89.2\%&	0.85&	91.1\%&	0.89&	74.6\%&	0.79&	73.7\%&	0.74&	68.9\%&	0.75\\
		\bottomrule
		\multicolumn{11}{l}{\tabincell{l}{* represents the GaussianNB algorithm. Nan means null value, and 0.0 indicates that no drifts have been \\detected.}}
	\end{tabular}
	\caption{Experiment results on toy datasets}
	\label{tab3}
\end{table*}

\begin{table*}[ht]
	\small
	\centering
	\begin{tabular}{lcccccccccc}
		\toprule
		&\multicolumn{2}{c}{Elec}&\multicolumn{2}{c}{Weather}&\multicolumn{2}{c}{Spam}&\multicolumn{2}{c}{Air}&\multicolumn{2}{c}{Poke}\\
		\toprule
		&Acc	&DN	&Acc&	DN	&Acc&	DN&	Acc&	DN&	Acc&	DN\\\toprule
		* &73.9\%&	0&	68.7\%&	0&	89.3\%&	0&	58.1\%&	0&	51.2\%	&0\\\midrule
		*+ADWIN	&79.1\%	&46	&69.5\%&	8&92.4\%&	3&	62.2\%&	70&	54.2\%&	1\\\midrule
		*+DDM&	82.1\%	&117&	70.3\%&	3	&91.9\%&	8&	61.9\%&	17&	54.2\%&	1\\\midrule
		*+EDDM&	82.6\%	&115&	73.7\%&	49	&93.4\%&	13&	61.5\%&	178	&54.2\%	&1\\\midrule
		*+HDDM-A&	83.0\%&	123&	72.3\%&	29	&92.1\%&	4&	62.6\%&	122	&54.5\%&	4\\\midrule
		*+HDDM-W&	82.7\%&	99&	72.6\%&	38	&92.2\%&	2&	61.5\%&	617	&52.0\%&	604\\\midrule
		*+Page-Hinkley&	78.2\%&	17&	69.6\%&	5	&91.8\%&	1&	62.8\%&	63&	54.1\%	&15\\\midrule
		*+KSWIN	&81.4\%&	91	&67.8\%&	45	&92.4\%&	5	&61.3\%&	664	&53.6\%&	306\\\midrule
		*+Meta-DD&	83.5\%&	125	&74.0\%&	50	&94.5\%	&12	&63.9\%&	155	&55.7\%	&267\\\midrule
		*+Meta-ADD& 85.7\%& 101 &76.1\%&    55  &95.2\% &11 &64.7\%&    136 &56.1\% &202\\
		\bottomrule
		\multicolumn{11}{l}{DN represents the number of detected drifts.}
	\end{tabular}
	\caption{Experiment results on real-world datasets}
	\label{tab5}
\end{table*}

{\bf Findings and Discussion.} The results of Experiment 3 are shown in Table \ref{tab3}. In general, the prediction accuracy of Meta-Add is better than the other drift detection methods over all data streams. In addition, Table \ref{tab3} also shows Meta-ADD has the largest F1 score, i.e., detecting concept drift by Meta-ADD is more authentic. Thus, our proposed Me-ta-ADD achieves better performance than current drift de-tection methods. A comparison of the two last lines of Table \ref{tab3} shows that the performance of Meta-ADD was improved by active learning. 

\subsection{Stream Learning on Real-world datasets}\label{sec43}

\subsubsection{Experiment 4. Drift Detection on Real-word Datasets}

This experiment investigates whether our proposed Meta-ADD continues to obtain the best performance on real-world concept drift evaluation datasets. We expect to see similar results to those of Experiment 3.

{\bf Experiment Settings.} The basic settings are the same as Experiment 3. The main difference is that we used a group of real-world datasets with unknown drift, and these were most often used to evaluate the performance of drift detection methods. These datasets were \cite{bifet2010moa}:
\begin{itemize}
\item {\bf Elec:} the Electricity Price Dataset has 45325 samples and each sample has 9 features
\item {\bf Weather:} the NAOO Weather Dataset has 18712 samples and each sample has 9 features.
\item {\bf Spam:} the Spam Filtering Dataset has 9829 samples and each sample has 501 features.
\item {\bf Air:} the Airline Delay Dataset has 539395 samples and each sample has 8 features.
\item {\bf Poke:} the Poker Hand Dataset has 923720 samples and each sample has 11 features.
\end{itemize}

In the Elec, Weather, Spam datasets, parameters $l$ and $n$ were set to 100 and 1, respectively. In the remaining datasets, parameters $l$ and $n$ were set to 100 and 25, respectively.

{\bf Findings and Discussion.} The learning accuracies of the different drift detection methods appear in Table \ref{tab5}. From Table \ref{tab5}, we can see Meta-ADD continues to obtain the best performance on all datasets. In addition, we compared the number of drifts detected using different settings of parameters $l$ and $n$. In general, with an increase in $l$ and $n$, the number of detected drifts decreases. For example, in the Elec dataset, when parameter $l = 50$, but $n$ increases from 1 to 25, the number of detected drifts decreases from 158 to 28.

\section{Conclusion}

In this paper, we propose a novel framework named Meta-ADD to learn a meta-detector for detecting concept drift. Different to the current drift detection methods, Meta-ADD does not rely on rules set by humans to judge whether concept drift has occurred, rather it proves that machines can be offline trained to classify concept drift, and thereby solve problems facing the existing current drift detection methods, i.e., cold start and an inability to classify concept drift. Hence, compared with current methods, using Meta-ADD to detect concept drift can better understand them. Extensive experimental results also show the effectiveness of Meta-ADD.

\bibliographystyle{named}
\bibliography{ijcai21}

\end{document}